\documentclass[]{alibaba-nlp}

\usepackage{amsmath, amsthm, amssymb}
\usepackage{bm}
\usepackage{pifont}
\usepackage{booktabs}
\usepackage{multirow}
\usepackage{tabularx}
\usepackage{makecell}
\usepackage{graphicx}
\usepackage{wrapfig}

\usepackage{enumitem}
\usepackage{xspace}
\usepackage{microtype}
\usepackage[table]{xcolor}

\usepackage{algorithm}
\usepackage{algorithmicx}
\usepackage{algpseudocode}

\usepackage{url}

\newcommand{\ours}{\textsc{UEmbed}\xspace}
\newcommand{\eg}{\hbox{\emph{e.g.,}}\xspace}
\newcommand{\ie}{\hbox{\emph{i.e.,}}\xspace}

\newcommand{\mmebv}{MMEB-v2\xspace}
\newcommand{\beir}{\textsc{BEIR}\xspace}

\title{UEmbed: Unified Sparse and Dense Multimodal Embeddings}

\author[1,2,3]{Tingyu Song*}
\author[2]{Mingxin Li*}
\author[2]{Yanzhao Zhang}
\author[2]{Dingkun Long}

\author[2]{Pengjun Xie}
\author[2]{Zhijie Nie}
\author[4]{Yilun Zhao}
\author[1,3]{Shu Wu}

\affiliation[1]{CASIA}
\affiliation[2]{Alibaba Group}
\affiliation[3]{University of Chinese Academy of Sciences}
\affiliation[4]{Yale University}
\contribution[*]{Equal Contributions.}
\abstract{
Sparse retrieval underpins modern search systems, from web search to retrieval-augmented generation. 
Existing work has introduced Learned Sparse Retrieval (LSR) to push beyond exact lexical matching toward richer semantics.
Yet LSR has so far remained tied to encoder-style bidirectional architectures, and its extension to multimodal settings still relies heavily on auxiliary cross-modal modules. 
To address these limitations, we introduce \ours (Unified Embedding), a decoder-only multimodal embedding model that produces both sparse lexical and dense representations in one causal forward pass. 
\ours appends $N$ learnable special tokens to the input and partitions the vocabulary into $N$ disjoint subsets. Each token's causal hidden state predicts sparse weights over its assigned subset, and the $N$ subsets are concatenated into the full sparse vector. 
Trained on public data, we release \ours at 2B, 4B, and 9B scales. 
\ours-9B scores 71.8 (dense) and 71.0 (sparse) on MMEB-v2, leading models trained on public data in dense retrieval and setting the state of the art for sparse retrieval. On BEIR, \ours also remains competitive with strong dense and sparse baselines. 
Furthermore, we demonstrate the practical utility of \ours across three dimensions: effectiveness, efficiency, and agentic applications. 
Overall, \ours offers a new paradigm: it unifies dense and sparse embeddings in one model, while further extending sparse retrieval to unify text and multimodal inputs. }
\contact{Tingyu Song (\email{songtingyu23@mails.ucas.ac.cn})}
\correspondingauthor{Yilun Zhao (\email{yilun.zhao@yale.edu}), Shu Wu (\email{shu.wu@nlpr.ia.ac.cn})}
\projectleader{Dingkun Long}
\projectpage{\url{https://alibaba-nlp.github.io/UEmbed}}
\date{\today}

\begin{document}

\maketitle

\begin{figure*}[!htbp]
\centering
\includegraphics[width=0.90\linewidth]{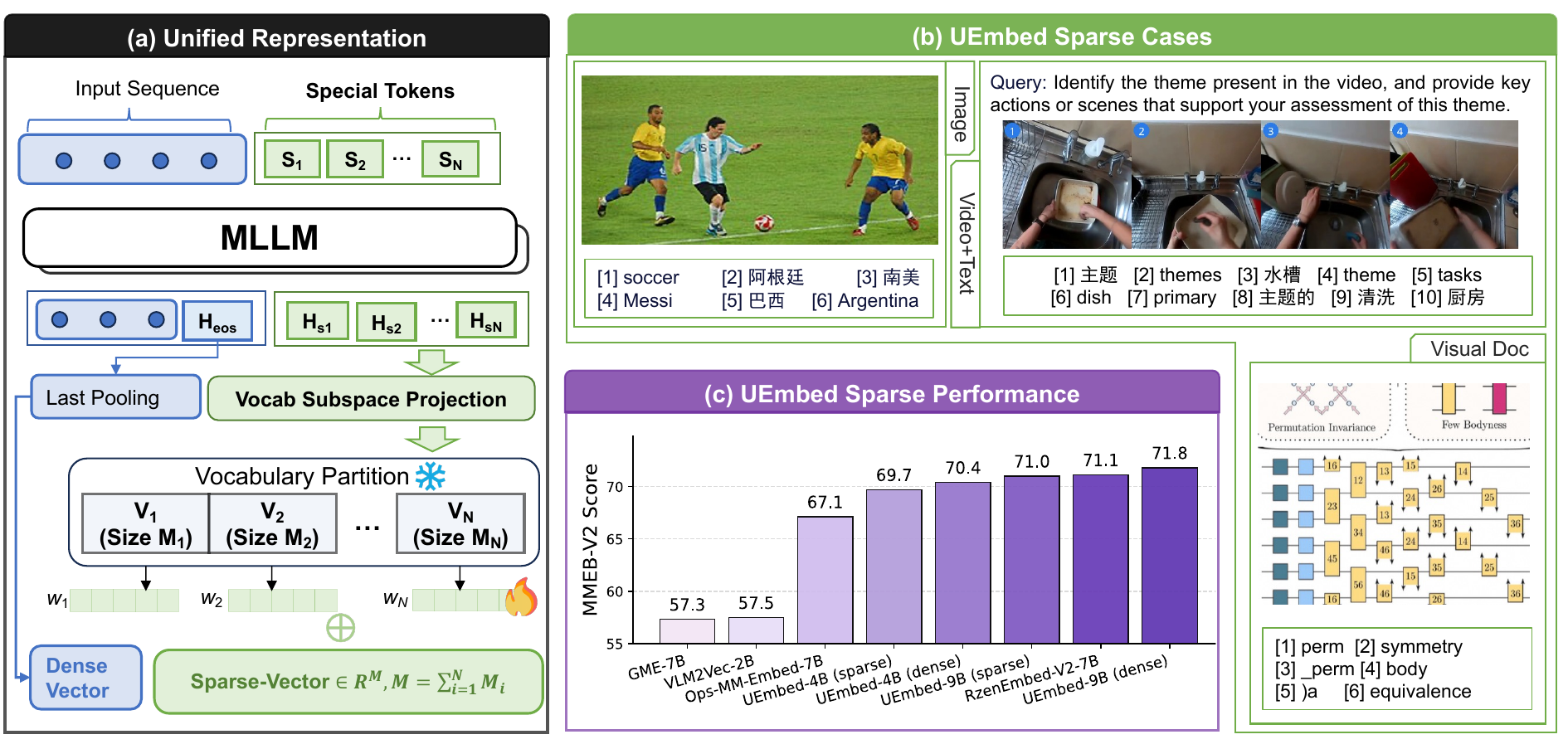}
\caption{Overview of \ours. (a) Framework: we append $N$ learnable special tokens to a decoder-only MLLM and assign each token a disjoint vocabulary subset. (b) Sparse cases: examples of activated lexical tokens produced by \ours-9B. (c) Performance: \ours achieves strong sparse and dense retrieval results on MMEB-v2.}
\label{fig:arxiv-figure1}
\end{figure*}

\section{Introduction}
\label{sec:intro}

Information retrieval (IR) underpins a broad range of real-world applications, including web search and question answering. 
Sparse lexical methods such as BM25~\citep{bm25} remain widely deployed for their efficiency and interpretability, but their reliance on exact term matching limits their ability to capture semantic similarity beyond surface-level lexical overlap.
Learned sparse retrieval (LSR) addresses this gap by training neural models to produce sparse lexical representations.
Representative methods such as SPLADE~\citep{spladev1, spladev2, spladepp} combine token-level transformations with max pooling, substantially improving retrieval quality over classical term-weighting schemes.

Despite this progress, LSR remains limited in three key respects.
(1) \emph{Architectural constraint:} current LSR methods are largely built on encoder-style, bidirectional architectures (\eg BERT~\citep{bert}). Due to the causal masking of unidirectional attention, decoder-only models cannot directly reuse traditional pooling strategies to capture the global context needed for sparse representations. 
Leveraging large language models (LLMs) for sparse retrieval has typically required either converting the causal backbone to a bidirectional one~\citep{bgem3} or adopting dedicated training curricula~\citep{laconic}, both of which incur additional training cost and break compatibility with optimized causal-model serving frameworks such as vLLM~\citep{vllm}.  
(2) \emph{Limited modality:} most LSR methods remain confined to text, as traditional approaches rely heavily on text-only BERT architectures that lack native multimodal capabilities. 
The few multimodal methods~\citep{visualsparta, stair} introduce auxiliary cross-modal modules rather than deriving sparse representations directly from a single multimodal backbone, making them difficult to extend to new modalities.
(3) \emph{Unexplored practical utility:} existing sparse methods are predominantly evaluated on traditional benchmarks such as COCO~\citep{coco}, leaving the practical advantages of sparse retrieval under-examined beyond accuracy. 

The primary challenge in deriving sparse representations from causal models is the information bottleneck: relying on a single token (\eg EOS, CLS) to project into a massive $|V|$-dimensional vocabulary space severely limits representational capacity. To overcome this, we present \ours (\textbf{\underline{U}}nified \textbf{\underline{Embed}}ding), which appends $N$ learnable special tokens to an MLLM and assigns each token a disjoint subset of the vocabulary. Specifically, these subsets are derived via $k$-means clustering, which explicitly encourages each token to represent a distinct spatial direction. Under causal attention, each special token summarizes the input from its position and predicts sparse weights over its assigned cluster. The $N$ subsets are then concatenated into the full sparse vector. This partitioned projection effectively circumvents the single-token representational bottleneck. The dense embedding is obtained from the hidden state of the EOS token preceding the special tokens.    
As shown in~\autoref{fig:arxiv-figure1}(b), \ours produces semantically meaningful lexical activations, indicating that the sparse representation captures cross-modal semantics.

We evaluate \ours on text and multimodal embedding benchmarks, with results summarized in~\autoref{fig:arxiv-figure1}(c).
On MMEB-v2, \ours-9B reaches 71.8 (dense) and 71.0 (sparse), closing the gap between the two modes to within one point at every scale we test.
On BEIR (9 datasets, nDCG@10), \ours remains competitive with strong dense and sparse baselines such as SPLADE-v3~\citep{spladev3}.
Moreover, our analysis surfaces three practical advantages of \ours. First, hybrid scoring over its dense and sparse modes consistently improves text and visual-document retrieval. Second, its design retains compatibility with high-throughput serving stacks (\eg vLLM) and inverted-index search. Finally, on BrowseComp-Plus~\citep{browsecompplus}, the sparse mode reduces tool-call costs while maintaining comparable recall. 

We summarize our contributions as follows:
\begin{itemize} [leftmargin=*]
\itemsep0em
    \item We propose \ours, which produces both sparse and dense retrieval representations from a single decoder-only backbone. 
    \item \ours unifies different modalities in one model and can be easily extended to diverse modality settings. To our knowledge, \ours is the first sparse model evaluated on MMEB-v2 and establishes a new state of the art in the sparse multimodal retrieval setting. 
    \item Our comprehensive evaluation demonstrates that the sparse mode of \ours improves both efficiency and effectiveness, while its strong performance in agentic search highlights its broad practical utility.
\end{itemize}

\section{Related Works}

\label{sec:related-works}

\subsection{Learned sparse retrieval}
Learned sparse retrieval replaces fixed lexical statistics, such as BM25, with neural models that both reweight observed terms and expand the representation with semantically related terms absent from the original text. SPLADE-based methods demonstrate that contextual token weighting and vocabulary expansion can make sparse retrieval competitive with strong dense baselines~\citep{spladev1, spladev2, spladev3}. Recent work extends this line of research using LLM backbones and stronger training recipes~\citep{laconic}. While related efforts either adapt decoder backbones toward bidirectional behavior for retrieval~\citep{laconic} or jointly train dense and sparse retrieval within one model~\citep{bgem3,sparsejoint}, our work differs by studying sparse retrieval natively under causal attention within a unified backbone. 

\subsection{Multimodal Retrieval}
Recently, Multimodal Large Language Models (MLLMs)~\citep{gemini2, qwen3vl2025, gpt5} have demonstrated exceptional performance across a wide range of tasks. Consequently, universal multimodal embedding models~\citep{ vlm2vec, thinkthenembed2025, qwenvlembedding} have been built upon these MLLM backbones, achieving highly competitive results on benchmarks such as \mmebv~\citep{vlm2vecv2}. However, multimodal sparse retrieval remains much less explored. Existing LSR work typically introduces auxiliary modules or an extra model to interpret non-text modalities~\citep{visualsparta, stair, blipsparse, frozen, splare}. Such auxiliary modules restrict the sparse model to a specific modality pair and complicate its extension to new modalities. Our work addresses this gap by combining decoder-native sparse retrieval, unified dense and sparse modeling, and multimodal support within a single causal backbone. 
\section{Method}

\autoref{fig:arxiv-figure1} illustrates the overall framework of \ours.
We first review preliminaries on contrastive learning and learned sparse retrieval in~\S\ref{sec:preliminaries}, then present the proposed method in~\S\ref{sec:method}, and finally describe the data curation in~\S\ref{sec:data}.

\subsection{Preliminaries}
\label{sec:preliminaries}

\paragraph{InfoNCE Loss.}
Given a query $q$, a positive document $d^{+}$, and a set of in-batch negatives $\{d^{-}_{j}\}_{j=1}^{K}$, the InfoNCE loss encourages the model to assign higher similarity to the positive pair:
\begin{equation}\label{eq:infonce}
\mathcal{L}_{\text{InfoNCE}} = -\log \frac{\exp\bigl(\text{sim}(q, d^{+}) / \tau\bigr)}{\sum_{d \in \{d^+\} \cup \mathcal{N}}\exp\bigl(\text{sim}(q, d) / \tau\bigr)},
\end{equation}
where $\mathcal{N}$ is the set of in-batch negatives, $\text{sim}(\cdot,\cdot)$ denotes a similarity function (\eg cosine similarity for dense retrieval, inner product for sparse retrieval), and $\tau$ is a temperature hyperparameter.

\paragraph{Learned Sparse Retrieval.}
SPLADE-based models~\citep{spladev1, spladev2} produce a sparse representation by projecting each token's hidden state onto the vocabulary and aggregating across the sequence with a max-pooling and saturating activation: 
\begin{equation}\label{eq:splade}
w_{t} = \max_{i=1}^{L} \log\bigl(1 + \text{ReLU}(\bm{h}_{i}^{\top} \bm{e}_{t})\bigr),
\end{equation}
where $\bm{h}_{i}$ is the hidden state of the $i$-th input token, $\bm{e}_{t}$ is the embedding of vocabulary term $t$, $L$ is the sequence length, and $w_{t}$ is the resulting weight for term $t$. The sparse vector $\bm{w} \in \mathbb{R}^{|V|}$ can be served by a standard inverted index.

\subsection{\ours Method}
\label{sec:method}

The key challenge in adapting sparse retrieval to decoder-only models is that unidirectional attention prevents each position from attending to future tokens, rendering max-pooling over all hidden states ineffective.
We address this by introducing $N$ special tokens appended to the input sequence and partitioning the vocabulary into $N$ disjoint subsets.

\paragraph{Compressing Vocabulary.}
LLM tokenizers often contain redundant tokens (\eg whitespace, accents). Before partitioning, we compress the vocabulary via accent stripping, lowercasing, and whitespace collapsing using NLTK~\citep{nltk}. Tokens sharing the same canonical form are merged into a single entry. For example,  Hello, HELLO, and héllò are all combined into hello. During scoring, we retain the maximum weight among them. This reduces the vocabulary size from 248,320 to 184,016. 

\paragraph{Partitioned Sparse Heads.}
Given an input sequence of length $L$, we append $N{=}16$ learnable special tokens $\langle s_1 \rangle, \dots, \langle s_N \rangle$ at the end. Each special token can attend to all preceding tokens, effectively summarizing the full input. 
To encourage each special token to capture a distinct semantic subspace, we employ k-means clustering to partition the vocabulary $V$ into $N$ disjoint subsets $V_1, \dots, V_N$ of approximately equal size. 

Each special token $\langle s_k \rangle$ is responsible for producing sparse weights over its assigned subset $V_k$ via a subset-specific sparse head. 
Let $H_{s_k}$ denote the final hidden state of $\langle s_k \rangle$. We compute the sparse weight for a vocabulary term $t \in V_k$ using a linear projection:
\begin{equation}\label{eq:subspace-sparse}
w_{t}^{(k)} = \log\bigl(1 + \text{ReLU}(\bm{W}_{t}^{(k)\top} \bm{h}_{s_k} + b_{t}^{(k)})\bigr),
\end{equation}
for all $t \in V_k$, where $\bm{W}_{t}^{(k)}$ and $b_{t}^{(k)}$ are the specific weight vector and bias scalar corresponding to term $t$ in the $k$-th sparse head. The sparse representation is obtained by concatenating all subset vectors:
\begin{equation}\label{eq:concat-sparse}
\bm{w} = \bigl[w_{t}^{(1)}\bigr]_{t \in V_1} \oplus \cdots \oplus \bigl[w_{t}^{(N)}\bigr]_{t \in V_N},
\end{equation}
where $\oplus$ denotes concatenation. 

\paragraph{Dense Representation.}
For dense retrieval, we use the hidden state of the EOS token preceding the special tokens as the dense embedding $\bm{d} \in \mathbb{R}^{D}$. When only dense retrieval is needed, the special tokens can be omitted entirely, incurring no extra forward computation.

\paragraph{Unified Objective.}
We train the model with a combined loss over both retrieval modes:
\begin{equation}\label{eq:total-loss}
\begin{split}
\mathcal{L} ={} & \mathcal{L}_{\text{InfoNCE}}^{\text{dense}} + \lambda\, \mathcal{L}_{\text{InfoNCE}}^{\text{sparse}}  + \alpha_{q}\, \mathcal{L}_{\text{FLOPS}}^{q} + \alpha_{d}\, \mathcal{L}_{\text{FLOPS}}^{d},
\end{split}
\end{equation}
where $\mathcal{L}_{\text{InfoNCE}}^{\text{dense}}$ and $\mathcal{L}_{\text{InfoNCE}}^{\text{sparse}}$ are computed using cosine similarity and inner product, respectively.
$\mathcal{L}_{\text{FLOPS}}^{q}$ and $\mathcal{L}_{\text{FLOPS}}^{d}$ are FLOPS regularizers~\citep{paria2020flops} that penalize the squared mean term weight across queries and documents respectively, encouraging sparsity.
$\lambda$, $\alpha_{q}$, and $\alpha_{d}$ are scalar coefficients balancing the four losses. The training details are provided in Appendix~\ref{sec:appendix-implementation}.

\subsection{Data Curation}\label{sec:data}

\paragraph{Data Sources.}
We draw training pairs from three publicly available datasets, comprising a total of 3.94M samples. 
(1)~\textbf{Echo-embedding training data}~\citep{echoemb}, which provides large-scale query-document pairs covering a wide range of domains;
(2)~\textbf{MLDR training data}~\citep{bgem3}, which provides the long document training data; and
(3)~\textbf{MMEB training sets}~\citep{vlm2vecv2}, which supply multimodal training data across diverse visual and cross-modal tasks. 

\paragraph{Hard Negative Mining.} 
As demonstrated by previous work~\citep{spladepp}, hard negatives are crucial for training learned sparse retrievers. 
Because the multimodal datasets lack the negatives required for effective training, we utilize Qwen3-VL-Embedding-8B~\citep{qwenvlembedding} as a teacher model to mine hard negatives for all multimodal data. 
Specifically, for each query, we use the teacher model to retrieve the top-$k$ most similar non-relevant documents from the corpus, treating these as our hard negatives. 

\section{Experiments}\label{sec:experiments}

\subsection{Experimental Setup}\label{sec:exp-setup}
\begin{table*}[htbp]
\centering
\renewcommand{\arraystretch}{1.1}
\setlength{\tabcolsep}{2.6pt}
\resizebox{\textwidth}{!}{
\begin{tabular}{l ccccc ccccc ccccc c}
\toprule[.1em]
\multirow{2}{*}{\textbf{Model}}
& \multicolumn{5}{c}{\textbf{Image}}
& \multicolumn{5}{c}{\textbf{Video}}
& \multicolumn{5}{c}{\textbf{VisDoc}}
& \multirow{2}{*}{\textbf{All}}\\
\cmidrule(lr){2-6} \cmidrule(lr){7-11} \cmidrule(lr){12-16}
& \textbf{CLS} & \textbf{QA} & \textbf{RET} & \textbf{GD} & \textbf{Avg.} & \textbf{CLS} & \textbf{QA} & \textbf{RET} & \textbf{MRET} & \textbf{Avg.} & \textbf{VDRv1} & \textbf{VDRv2} & \textbf{VR} & \textbf{OOD} & \textbf{Avg.} & \\
\midrule
\textbf{\# of Datasets}
& 10 & 10 & 12 & 4 & 36 & 5 & 5 & 5 & 3 & 18 & 10 & 4 & 6 & 4 & 24 & 78 \\
\midrule
Qwen3-VL-Embedding-8B & 74.2 & 81.1 & 80.2 & 92.3 & 80.1 & 78.4 & 71.0 & 58.7 & 56.1 & 67.2 & 87.2 & 69.9 & 88.7 & 73.3 & 82.4 & 77.8 \\
Qwen3-VL-Embedding-2B & 70.3 & 74.3 & 74.8 & 88.5 & 75.0 & 71.9 & 64.9 & 53.9 & 53.3 & 61.9 & 84.4 & 65.3 & 86.4 & 69.4 & 79.2 & 73.2 \\
RzenEmbed-V2-7B & \textbf{70.6} & 71.7 & \textbf{78.5} & \textbf{92.1} & \textbf{75.9} & 58.8 & 63.5 & 51.0 & 45.5 & 55.7 & \textbf{89.7} & 60.7 & \textbf{88.7} & 34.7 & 75.5 & 71.1 \\
Embed-RL-4B & 63.7 & 70.5 & 71.3 & 91.3 & 71.2 & 57.6 & 58.4 & 45.1 & 49.5 & 53.0 & 80.2 & 53.4 & 84.9 & 67.1 & 74.7 & 68.1 \\
Embed-RL-2B & 62.8 & 67.9 & 68.6 & 90.4 & 69.2 & 57.0 & 55.9 & 45.1 & 49.4 & 52.1 & 80.0 & 52.0 & 84.6 & 65.7 & 74.1 & 66.8 \\
Ops-MM-Embed-7B & 69.7 & 69.6 & 73.1 & 87.2 & 72.7 & 59.7 & 62.2 & 45.7 & 43.2 & 53.8 & 80.0 & 59.6 & 79.3 & 33.7 & 68.7 & 67.1 \\
Ops-MM-Embed-2B & 68.1 & 65.1 & 69.2 & 80.8 & 69.0 & 53.6 & 55.6 & 41.8 & 33.7 & 47.6 & 76.4 & 53.2 & 77.6 & 32.5 & 65.5 & 63.0 \\
UniME-7B & 67.1 & 69.2 & 71.9 & 84.8 & 71.2 & 48.6 & 60.7 & 38.2 & 39.3 & 47.5 & 75.7 & 50.5 & 83.7 & 29.1 & 65.7 & 64.1 \\
UniME-2B & 64.8 & 62.8 & 67.6 & 77.2 & 66.6 & 44.3 & 51.0 & 32.9 & 39.7 & 42.2 & 72.4 & 46.2 & 79.2 & 28.9 & 62.5 & 59.7 \\
GME-7B & 57.6 & 34.6 & 71.2 & 59.5 & 55.9 & 37.3 & 50.3 & 28.3 & 37.5 & 38.4 & 89.5 & 55.5 & 85.0 & 34.7 & 73.6 & 57.3 \\
VLM2Vec-V2-2.0B & 62.9 & 56.4 & 69.6 & 77.1 & 64.9 & 39.2 & 34.7 & 28.4 & 37.5 & 34.7 & 74.4 & 44.6 & 79.3 & 31.2 & 63.5 & 57.5 \\
\midrule
\ours-2B (dense) & 59.2 & 67.9 & 68.8 & 85.6 & 67.8 & 56.3 & 52.8 & 45.8 & 41.8 & 50.0 & 82.3 & 59.9 & 85.6 & 67.9 & 77.0 & 66.5 \\
\ours-2B (sparse) & 58.9 & 66.1 & 68.5 & 85.1 & 67.0 & 53.7 & 49.5 & 44.2 & 40.3 & 47.7 & 82.3 & 60.6 & 84.5 & 67.3 & 76.7 & 65.5 \\
\ours-4B (dense) & 62.3 & 72.3 & 72.6 & 88.4 & 71.4 & 61.4 & 65.0 & 50.1 & 47.9 & 57.0 & 84.1 & \textbf{61.0} & 87.3 & \textbf{70.4} & 78.8 & 70.4 \\
\ours-4B (sparse) & 61.9 & 70.2 & 72.3 & 88.0 & 70.6 & 61.8 & 63.3 & 48.3 & 47.1 & 56.0 & 84.3 & 60.6 & 87.2 & 69.5 & 78.6 & 69.7 \\
\ours-9B (dense) & 64.7 & \textbf{73.9} & 73.9 & 90.6 & 73.2 & \textbf{63.0} & \textbf{65.6} & \textbf{51.8} & \textbf{53.5} & \textbf{59.0} & 85.5 & 59.1 & 87.9 & \textbf{70.4} & \textbf{79.2} & \textbf{71.8} \\
\ours-9B (sparse) & 64.3 & 72.3 & 73.7 & 89.5 & 72.5 & 60.9 & 63.7 & 50.9 & 52.4 & 57.5 & 86.2 & 57.9 & 87.6 & 69.7 & 79.1 & 71.0 \\
\bottomrule[.1em]
\end{tabular}
}
\caption{Results on \mmebv. \textbf{Bold} marks the best score in each column among models (\ie excluding Qwen3-VL-Embedding, which is trained with large-scale proprietary data).}
\label{tab:main_results}
\end{table*}

\paragraph{Backbones and Model Sizes.}
We instantiate \ours from the Qwen3.5 family at three scales: 2B, 4B, and 9B parameters.
For each scale, we train a single checkpoint that supports both dense and sparse retrieval. 

\paragraph{Multimodal Benchmark and Baselines.}
We evaluate \ours on \mmebv~\citep{vlm2vecv2}, which provides a comprehensive evaluation for multimodal embedding tasks. 
To comprehensively position our model within the current landscape, we compare it against a diverse selection of current strong baselines:
(1) Qwen3-VL-Embedding~\citep{qwenvlembedding}, a strong vision-language embedding baseline trained with large-scale data; 
(2) RzenEmbed~\citep{rzenembed}, for which we adopt the competitive RzenEmbed-V2-7B variant; 
(3) Embed-RL~\citep{embedrl}, for which both the 2B and 4B versions are selected; 
(4) Ops-MM-Embed~\citep{ops} and (5) UniME~\citep{unime2025}, where we select both the 2B and 7B versions for a multi-scale comprehensive comparison; 
(6) GME-7B~\citep{gme}, a robust 7B multimodal retriever; and
(7) VLM2Vec~\citep{vlm2vecv2}, specifically evaluating the VLM2Vec-V2-2.0B version.

\paragraph{Text Benchmark and Baselines.}
We evaluate \ours on nine datasets from \beir and adopt nDCG@10 as the evaluation metric. Details of the evaluation setup are provided in Appendix~\ref{sec:appendix-experiment-details}.
To ensure a fair comparison, we specifically select multimodal embedding models as our dense baselines, intentionally excluding purely text-focused models. Furthermore, we incorporate competitive sparse models to establish a comprehensive evaluation framework. Our selected baselines include:
(1) GME-7B~\citep{gme}, a robust 7B multimodal embedding model that demonstrates strong text retrieval performance on BEIR; 
(2) Qwen3-VL-Embedding~\citep{qwenvlembedding}, a powerful vision-language embedding baseline trained on large-scale data that also exhibits competitive text-only retrieval capabilities;
(3) SPLADE-v3~\citep{spladev3}, representing the latest iteration of the classical sparse retrieval method; and
(4) Echo-Mistral-SPLADE~\citep{mistralsplade}, which utilizes a powerful LLM backbone to generate highly effective sparse representations.

\subsection{Multimodal Results}\label{sec:multimodal-results}

\autoref{tab:main_results} reports MMEB-v2 results across all three super-categories. 

\paragraph{\ours is competitive with the strongest open multimodal embedders.}
\ours-9B (dense) reaches an aggregate score of 71.8. While this is surpassed by Qwen3-VL-Embedding-8B, it is crucial to note that Qwen's models benefit from multi-stage training with vast, proprietary datasets, making a direct comparison challenging. When compared with other models that have publicly available checkpoints and are trained on open datasets, \ours-9B (dense) leads models trained on publicly available data, outperforming models like RzenEmbed-V2-7B (71.1) and Ops-MM-Embed-7B (67.1). At smaller scales, the picture is consistent: \ours-4B (dense) with a score of 70.4 surpasses every other model in the 4B-parameter class, such as Embed-RL-4B (68.1). Furthermore, our most compact model, \ours-2B (dense) at 66.5, is highly competitive with, and even outperforms some, 7B-scale open baselines like UniME-7B (64.1). To better understand the gap between \ours and Qwen3-VL-Embedding, we also conduct an analysis in Appendix~\ref{sec:appendix-case-study}. 

\paragraph{Sparse retrieval remains competitive with dense retrieval in the multimodal regime.}
Our results demonstrate that sparse embeddings are a viable and powerful alternative to dense ones in the multimodal domain. As shown in the table, the performance of our sparse models is remarkably close to their dense counterparts across all scales, with the overall performance gap being at most 1.0 point (\eg 71.8 vs. 71.0 for the 9B model). To our knowledge, these are the first reported results for sparse multimodal embeddings on this benchmark, and they significantly outperform established dense models. For instance, \ours-4B (sparse) at 69.7 surpasses the dense Ops-MM-Embed-7B (67.1). A closer look reveals that sparse models are particularly effective on Visually-rich Document (VisDoc) tasks, where the performance drop is minimal (\eg 79.2 vs. 79.1 for the 9B model). This suggests that the inherent structure of sparse embeddings may be well-suited for document-based tasks, while the performance gap is slightly more pronounced in the Video category. 

\begin{table*}[htbp]
\centering
\resizebox{\textwidth}{!}{
\begin{tabular}{l c c c c c c c c c c }
\toprule[.1em]
\textbf{Model} & \textbf{ArguAna} & \makecell[c]{\textbf{FiQA}\\\textbf{2018}} & \textbf{NFCorpus} & \textbf{NQ} & \textbf{Quora} & \textbf{SCIDOCS} & \textbf{SciFact} & \textbf{COVID} & \makecell[c]{\textbf{Touche}\\\textbf{2020}} & \textbf{Avg} \\
\midrule
\multicolumn{11}{c}{\emph{Dense Models}} \\
\midrule
GME-7B & \textbf{64.6} & \textbf{57.1} & 38.4 & \textbf{67.7} & 88.1 & \textbf{27.4} & 62.3 & 52.6 & 23.3 & 53.5 \\
Qwen3-VL-Embedding-2B & 43.2 & 41.8 & 37.2 & 58.1 & 86.6 & 21.9 & 75.1 & \textbf{89.6} & \textbf{30.8} & 53.8 \\
Qwen3-VL-Embedding-8B & 45.8 & 47.8 & 39.7 & 64.5 & 86.5 & 24.9 & \textbf{79.4} & 84.1 & 26.9 & 55.5 \\
\ours-2B & 58.1 & 47.5 & 37.8 & 55.5 & 89.3 & 20.3 & 75.3 & 80.2 & 18.2 & 53.6 \\
\ours-4B & 60.0 & 53.4 & \textbf{40.5} & 60.9 & 89.2 & 22.5 & 77.0 & 82.7 & 18.1 & 56.0 \\
\ours-9B & 62.5 & 54.3 & 39.8 & 61.9 & \textbf{89.9} & 23.4 & 77.8 & 77.9 & 19.2 & \textbf{56.3} \\
\midrule
\multicolumn{11}{c}{\emph{Sparse Models}} \\
\midrule
SPLADE-v3 & 50.9 & 37.4 & 35.7 & 58.6 & 81.4 & 15.8 & 71.0 & 74.8 & \textbf{29.3} & 50.5 \\
Echo-Mistral-SPLADE & 56.2 & \textbf{57.7} & \textbf{42.3} & 56.0 & 86.7 & \textbf{25.6} & \textbf{77.2} & 76.8 & 18.0 & \textbf{55.2} \\
\ours-2B & 55.7 & 45.1 & 35.9 & 53.4 & 88.2 & 18.0 & 70.8 & \textbf{84.8} & 18.7 & 52.3 \\
\ours-4B & \textbf{58.8} & 49.4 & 38.4 & \textbf{59.9} & 88.6 & 20.8 & 74.4 & 75.4 & 16.7 & 53.6 \\
\ours-9B & 58.4 & 51.2 & 38.1 & \textbf{59.9} & \textbf{88.7} & 20.6 & 74.5 & 82.1 & 23.4 & \textbf{55.2} \\
\bottomrule[.1em]
\end{tabular}
}
\caption{nDCG@10 on BEIR datasets. \textbf{Bold} marks the best score in each column within each retrieval mode.}
\label{tab:beir}
\end{table*}
\subsection{Text Results}\label{sec:text-results}
In the dense retrieval setting, as shown in~\autoref{tab:beir}, \ours demonstrates highly competitive performance across the evaluated benchmarks. Our \ours-9B model achieves the highest average nDCG@10 score of \textbf{56.3} among the compared models, with \ours-4B closely following at 56.0. This places our models ahead of strong, recent baselines like Qwen3-VL-Embedding-8B and GME-7B. Furthermore, \ours shows particular strength on specific datasets, such as achieving 89.9 on Quora (\ours-9B) and 40.5 on NFCorpus (\ours-4B).  

In the sparse retrieval setting, \ours-9B reaches an average of \textbf{55.2}, effectively matching the strong specialist model Echo-Mistral-SPLADE (55.2). Crucially, \ours achieves this parity while simultaneously supporting dense retrieval and multimodal inputs within a single backbone. These results position \ours as a unified model that sacrifices little text performance while gaining robust multimodal and dense capabilities. 

\section{Analysis}\label{sec:analysis}

To better understand \ours, this section analyzes the model from the following perspectives: 
(1)~\textbf{Component Analysis:} controlled ablations to validate the contribution of individual design choices (\S\ref{sec:ablation});
(2)~\textbf{Cross-modal Robustness:} case studies illustrating the model's performance across diverse modalities (\S\ref{sec:case-study}); and
(3)~\textbf{Practical Advantages:} the utility of sparse retrieval in hybrid scoring, serving efficiency, and agentic search.

\subsection{Component Analysis} \label{sec:ablation}

Unless otherwise noted, ablations in this section use the 2B backbone trained on a random 500k-instance subset from the full mixture, and are evaluated on MMEB-v1~\citep{vlm2vec}. 
\paragraph{Baseline Comparison.}
\begin{table}[htbp]
\centering
\begin{minipage}[t]{0.49\textwidth}
\vspace{0pt}
\centering
\footnotesize
\setlength{\tabcolsep}{3pt}
\begin{tabular}{lccccc}
\toprule
Model & CLS & QA & RET & GRD & Avg. \\
\midrule
SPLADE & 55.8 & 57.1 & 62.9 & 80.9 & 61.3 \\
\ours (dense) & 55.5 & 65.3 & 65.1 & 83.3 & 64.5 \\
\ours (sparse) & 54.7 & 63.1 & 64.8 & 81.8 & 63.4 \\
\bottomrule
\end{tabular}
\captionof{table}{Comparison with the bidirectional SPLADE baseline on the image subset of MMEB-v1.}
\label{tab:baseline-compare}
\end{minipage}
\hfill
\begin{minipage}[t]{0.49\textwidth}
\vspace{0pt}
\centering
\footnotesize
\setlength{\tabcolsep}{3pt}
\begin{tabular}{lccccc}
\toprule
Partition & CLS & QA & RET & GRD & Avg. \\
\midrule
Random & 55.0 & 62.3 & 64.2 & 81.0 & 63.0 \\
Max-distance & 55.1 & 62.4 & 64.9 & 79.8 & 63.2 \\
Semantic (ours) & 54.7 & 63.1 & 64.8 & 81.8 & 63.4 \\
\bottomrule
\end{tabular}
\captionof{table}{Sparse performance under different vocabulary partitioning strategies on the image subset of MMEB-v1.}
\label{tab:partition-compare}
\end{minipage}
\end{table}

To assess whether the gains in~\S\ref{sec:multimodal-results} are attributable to the \ours recipe, we compare \ours-2B against a SPLADE-based baseline. This baseline shares the same Qwen3.5 backbone, training data, and FLOPS regularization, but adopts bidirectional attention with a standard SPLADE max-pooling head (\autoref{eq:splade}). \autoref{tab:baseline-compare} demonstrates that \ours-2B surpasses this bidirectional baseline in \emph{both} modes: $+3.2$ points for dense (61.3 $\rightarrow$ 64.5) and $+2.1$ for sparse (61.3 $\rightarrow$ 63.4) on average. The margins are most pronounced on IMG-QA ($+8.2$ dense, $+6.0$ sparse), suggesting the standard SPLADE recipe underutilizes the autoregressive backbone's inherent QA capabilities. Thus, our unified causal formulation is intrinsically beneficial to embedding quality, beyond merely offering serving convenience. 

\paragraph{Robustness to Joint Training.}
\begin{figure*}[!t]
\centering
\begin{subfigure}[t]{0.30\linewidth}
    \centering
    \includegraphics[width=\linewidth]{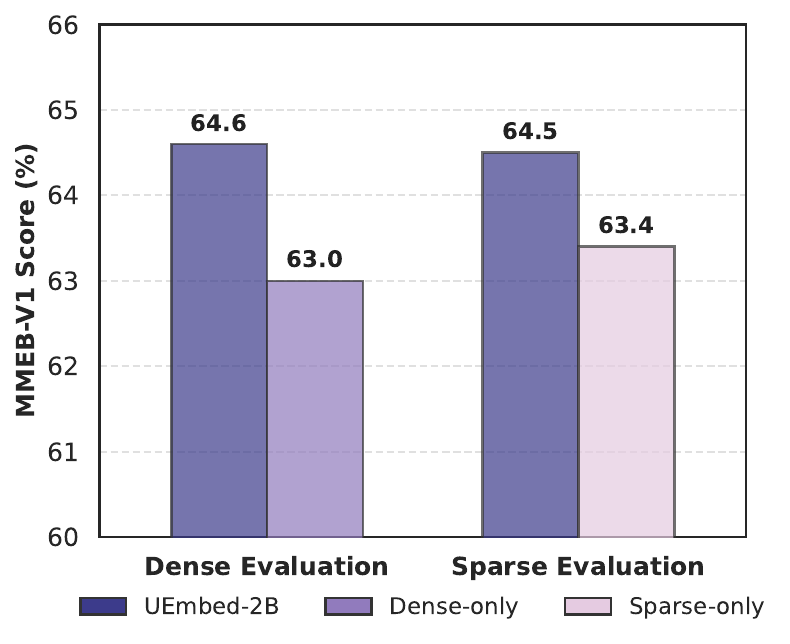}
    \caption{Joint vs.\ single-mode training.}
    \label{fig:dense-sparse-mode-bar}
\end{subfigure}
\hfill
\begin{subfigure}[t]{0.30\linewidth}
    \centering
    \includegraphics[width=\linewidth]{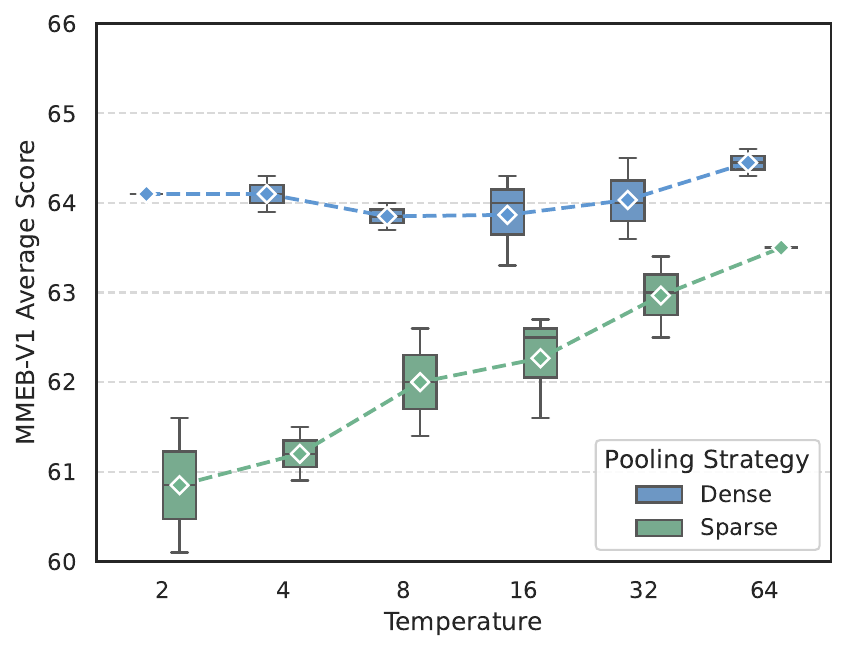}
    \caption{Sparse temperature $\tau_s$.}
    \label{fig:temp-ablation-boxplot}
\end{subfigure}
\hfill
\begin{subfigure}[t]{0.30\linewidth}
    \centering
    \includegraphics[width=\linewidth]{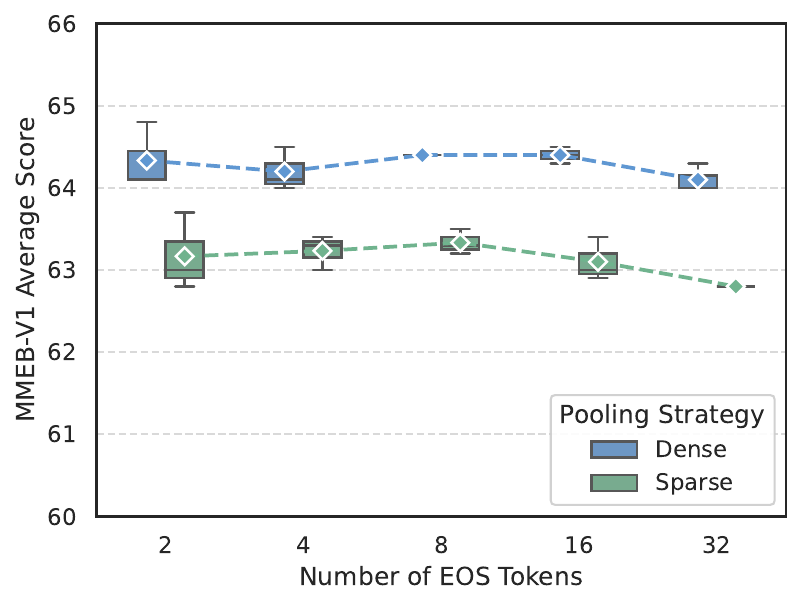}
    \caption{Number of special tokens $N$.}
    \label{fig:nes-ablation-boxplot}
\end{subfigure}
\caption{Ablation studies of \ours: (a) joint training matches single-mode specialists; (b)~sweeping the sparse softmax temperature $\tau_s$; and (c)~sweeping the number of appended special tokens $N$.}
\label{fig:ablation-panels}
\end{figure*}

A common pitfall in multi-task learning is negative transfer between objectives. To verify if our joint objective (\autoref{eq:total-loss}) compromises either retrieval mode, we train dense-only ($\lambda{=}0$) and sparse-only baselines under identical conditions. As shown in~\autoref{fig:dense-sparse-mode-bar}, the jointly trained \ours model closely matches the performance of both single-mode specialists. This confirms that the dual-mode capability incurs a negligible performance penalty during training. 

\paragraph{Vocabulary Partitioning.}
The sparse head partitions the vocabulary $V$ into $N$ disjoint subsets. We compare three balanced partitioning strategies: (1)~\textbf{Random} uniform assignment; (2)~\textbf{Max-distance}, which pairs maximally distant $k$-means clusters to enforce semantic diversity within subsets; and (3)~\textbf{Ours (Semantic clustering)}, which assigns individual $k$-means clusters to distinct subsets. Semantic clustering yields the best performance, while random partitioning performs worst. This suggests that grouping semantically related tokens allows each special token to act as a ``soft topic specialist,'' optimizing its limited representational capacity rather than struggling to model unrelated concepts simultaneously.

\paragraph{Hyperparameter Sensitivity.}
We sweep two hyperparameters specific to the sparse head:
\textit{Sparse temperature ($\tau_s$).} Unlike the dense head's $[-1,1]$ cosine similarities, sparse inner products have a much larger dynamic range. A shared temperature would therefore over-sharpen the sparse softmax. \autoref{fig:temp-ablation-boxplot} demonstrates that decoupling the temperatures and using a larger $\tau_s$ improves sparse performance without degrading dense retrieval; we use $\tau_s{=}32$ as our default, noting that $\tau_s{=}64$ performs comparably in the single-seed sweeps we report. 
\textit{Number of special tokens ($N$).} Sweeping $N \in \{2,4,8,16,32\}$ (\autoref{fig:nes-ablation-boxplot}) reveals stable performance up to $N{=}16$, with a clear drop at $N{=}32$. We attribute this degradation to overly small vocabulary subsets and inflated contrastive sequence lengths. We adopt $N{=}16$ as our implementation default. 

\subsection{Case Study} \label{sec:case-study}
\begin{figure}[t]
\centering
\includegraphics[width=0.90\linewidth]{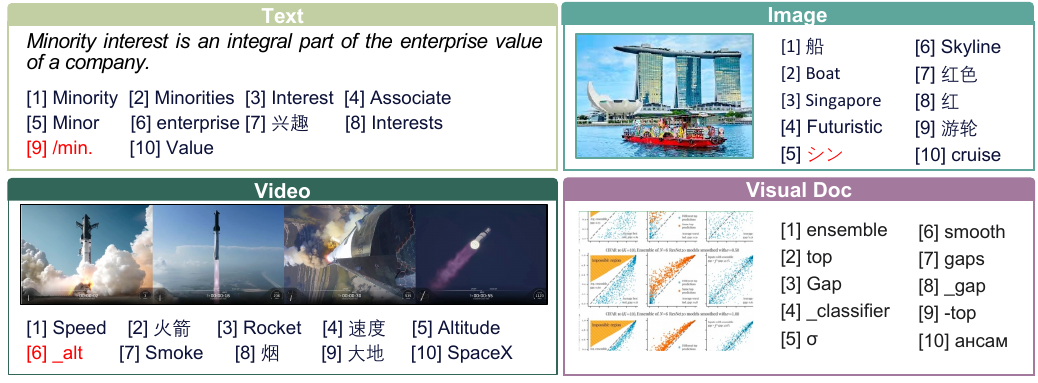}
\caption{Case study of \ours on different modalities. We visualize the top-10 activated sparse tokens.}
\label{fig:case-study}
\end{figure}

Building upon the examples in~\autoref{fig:arxiv-figure1}(b), we present additional case studies across various modalities. As illustrated in~\autoref{fig:case-study}, our model demonstrates a robust capacity to identify essential components within the input and activate semantically relevant concepts. For instance, it can deduce an image's location as Singapore based on its skyline and accurately identify a rocket as belonging to SpaceX.

However, we also observe two primary limitations:
(1) \emph{Generation of Anomalous Tokens}: The model occasionally produces non-standard tokens (\eg ``\_alt''). Unlike traditional masked language models such as BERT, which operate within a restricted and highly structured vocabulary, the exceptionally large vocabulary size of modern LLMs increases the probability of generating such edge-case tokens. This suggests that further vocabulary compression or targeted pruning may be necessary to ensure representational stability. 
(2) \emph{Limited Multilingual Support}: The model predominantly activates English and Chinese tokens while neglecting other languages. This bias primarily stems from our training corpus, which is heavily skewed toward English and Chinese, as well as the inherent cross-lingual limitations of the underlying foundation model.

\subsection{Practical Advantages}

Sparse retrieval offers structural advantages in real-world deployments. In this section, we examine three aspects of \ours: (1) the \emph{effectiveness} gain from combining its sparse and dense modes via hybrid scoring; (2) the deployment-side \emph{efficiency} enabled by autoregressive serving and native inverted-index compatibility; and (3) its utility as a retriever in downstream \emph{agentic search} workflows.

\paragraph{Effectiveness.} \label{sec:effectiveness}

Because \ours produces sparse and dense representations in a single forward pass, the two modes can be combined without an additional encoding pass. We construct a hybrid score by linearly interpolating the dense and sparse similarities; the detailed formulation and per-modality weights are provided in Appendix~\ref{sec:appendix-ablation-study}. 
\autoref{tab:effective-analysis} shows hybrid scoring boosts \emph{Text} ($+0.3$) and \emph{VisDoc} ($+0.5$), where exact lexical matching complements dense semantics. In contrast, natural images and video frames carry little surface-form lexical information, leaving sparse retrieval with limited additional signal to contribute. Although sparse retrieval provides potential gains on lexically-dependent tasks, the aggregate improvement appears modest due to dilution across the diverse \mmebv and \beir benchmarks.

\paragraph{Efficiency.} \label{sec:efficiency}

\begin{wraptable}{r}{0.45\linewidth}
\centering
\small
\vspace{-1.2em}
\setlength{\tabcolsep}{4pt}
\begin{tabular}{lcccc}
\toprule[.1em]
Model & TXT & IMG & VID & VDR \\
\midrule
\ours (dense)  & 53.6 & 67.8 & 50.0 & 77.0 \\
\ours (sparse) & 52.3 & 67.0 & 47.7 & 76.7 \\
\ours (hybrid) & 53.9 & 67.9 & 50.0 & 77.5 \\
\bottomrule[.1em]
\end{tabular}
\caption{Modality-wise performance of \ours-2B under dense, sparse, and hybrid modes.}
\label{tab:effective-analysis}
\vspace{-1em}
\end{wraptable}

\ours offers two deployment-side advantages. First, because the model is purely autoregressive, embedding generation is directly compatible with high-throughput serving stacks such as vLLM~\citep{vllm}. Second, the sparse representations plug into standard inverted indices, enabling scalable lexical retrieval in production environments. We defer a detailed analysis of inverted-index search efficiency to Appendix~\ref{sec:appendix-ablation-study}. 

\paragraph{Application: Agentic Search.} \label{sec:application}
\begin{table}[htbp]
\centering
\begin{tabular}{lcccc}
\toprule[.1em]
Model & Accuracy (\%) $\uparrow$ & Recall (\%) $\uparrow$ & Avg. Search $\downarrow$ & Calibration Error (\%) \\
\midrule
BM25 & 36.87 & 43.02 & 17.91 & -- \\
Qwen3-Embedding-8B & 44.46 & 62.32 & 30.37 & -- \\
Qwen3-VL-Embedding-2B & 26.87 & 38.20 & 34.99 & 12.13 \\
Qwen3-VL-Embedding-8B & 31.45 & 44.42 & 34.19 & 12.35 \\
\midrule
\ours-2B (dense) & 44.10 & 53.47 & 39.19 & 16.69 \\
\ours-2B (sparse) & 44.05 & 57.04 & 32.67 & 8.54 \\
\ours-4B (dense) & 51.57 & 61.03 & 38.47 & 14.34 \\
\ours-4B (sparse) & 45.54 & 60.10 & 31.85 & 8.79 \\
\ours-9B (dense) & 49.76 & 64.72 & 33.68 & 7.06 \\
\ours-9B (sparse) & 49.76 & 64.33 & 31.05 & 8.16 \\
\bottomrule[.1em]
\end{tabular}
\caption{Results on BrowseComp-Plus. ``Avg. Search'' denotes the average number of search rounds per question.  }
\label{tab:agent-app}
\end{table} 
As noted by \citet{meng2026revisiting}, LLM agents frequently issue short, keyword-dense queries, a regime where sparse retrieval typically outperforms dense alternatives. 
To further explore this problem, we test \ours on BrowseComp-Plus~\citep{browsecompplus}, a benchmark where an LLM uses retrieval as a tool during iterative reasoning. We utilize DeepResearch-30A3B~\citep{tongyidr} as the backbone reasoning engine. As shown in~\autoref{tab:agent-app}, the sparse mode of \ours consistently requires fewer tool-call search rounds than its dense counterpart. Recall is higher at the 2B scale and comparable at 4B and 9B. These results validate sparse retrieval as a practical choice for cost-sensitive, iterative reasoning loops where keyword-dense queries dominate. 
\section{Conclusion}

In this work, we present \ours, a decoder-only multimodal embedding model that natively unifies dense and sparse retrieval within a single causal forward pass. By overcoming the inherent bottlenecks of causal attention, \ours systematically eliminates the reliance on bidirectional encoders and auxiliary cross-modal modules. Our comprehensive evaluations demonstrate that \ours establishes a new state of the art in sparse multimodal retrieval while remaining highly competitive in dense retrieval. Through rigorous component analysis, we validate our algorithmic designs and provide practical settings. Crucially, \ours delivers strong effectiveness together with native compatibility with autoregressive serving stacks and inverted indices, while demonstrating cost advantages in agentic search scenarios. Overall, it transforms sparse retrieval from a legacy standalone module into a native byproduct of MLLMs.  
\section*{Limitations}

While \ours establishes a strong foundation for unified multimodal retrieval, we identify several areas for future improvement. 
(1) \emph{Language and Cultural Bias:} The training corpus utilized in this work is predominantly skewed toward English and Chinese. Consequently, the sparse head activations exhibit limited cross-lingual generalization. Extending the framework's sparse capabilities to a broader range of languages will necessitate highly language-diverse, large-scale training data. 
(2) \emph{Vocabulary Stability and Artifacts:} Unlike traditional masked language models with constrained vocabularies, operating over the expansive vocabulary of modern LLMs occasionally leads to anomalous token activations (\eg non-standard subwords or artifacts like ``\_alt''). Future iterations can explore targeted vocabulary pruning, refined semantic clustering, or post-hoc filtering mechanisms to ensure representational stability in strict production environments. 
(3) \emph{Modality-Specific Performance Gaps:} While the sparse and dense representations maintain close performance parity on text and static visual documents, we observe a slightly more pronounced gap in the video domain. We attribute this to the high information density and temporal dynamics inherent to video frames. This suggests that naively pooling spatiotemporal data into a flat sparse vector might encounter capacity bottlenecks. 
\section*{Ethical Consideration}

Our model is trained exclusively on publicly available datasets, and we release model weights to facilitate reproducibility.
We acknowledge that embedding models can inherit biases present in their training data and backbone LLM; practitioners should evaluate fairness across demographic groups before deployment in high-stakes retrieval applications.
The sparse representations provide a degree of interpretability by surfacing activated vocabulary terms, which may aid bias auditing compared to purely dense models. 

\bibliographystyle{assets/plainnat}
\bibliography{custom,llm}

\appendix

\clearpage
\section{Implementation Details}\label{sec:appendix-implementation}

\subsection{Training Data} \label{sec:appendix-data}

\begin{wrapfigure}{r}{0.4\linewidth}
\centering
\includegraphics[width=\linewidth]{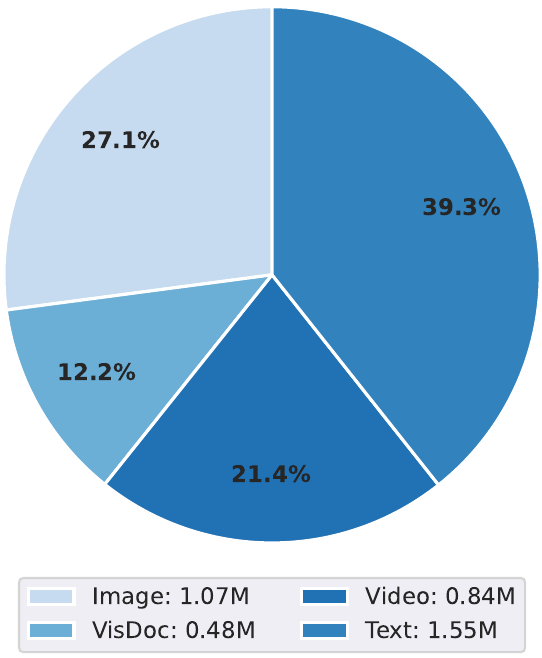}
\caption{Distribution of training data sources.}
\label{fig:training-data-pie}
\end{wrapfigure}

We curate a multi-source training mixture that covers text, image, video, and visually-rich document retrieval.
The composition ratio for each data source is shown in~\autoref{fig:training-data-pie}, and per-dataset retrieval instructions are listed in~\autoref{tab:train_instructions_text} and~\autoref{tab:train_instructions_multimodal}.

\paragraph{Text Training Data.}
To enable text embedding capabilities, we include training pairs from Echo-Embedding~\citep{echoemb} and M3-Embedding~\citep{bgem3}.
The text subset comprises NLI, DuReader, ELI5, FEVER, HotpotQA, MIRACL, MrTyDi, MSMARCO Passage, MSMARCO Document, Natural Questions, Quora Duplicates, SQuAD, T2Ranking, TriviaQA, and MLDR.
For standard text tasks we set a maximum sequence length of 1{,}500 tokens; for long-document retrieval (MLDR) we extend this to 1{,}800 tokens.

\paragraph{Multimodal Training Data.}
For video retrieval, we utilize VideoCaption300k and VideoQA240k from LLaVA-Hound~\citep{llava-hound}.
We sample 8 frames per video and allocate up to 200 tokens per frame.
For visual document retrieval, we incorporate ColPali train (118k), VisRAG-Synthetic (239k), and VisRAG-IID (123k) from ViDoRe~\citep{vidore} and VisRAG~\citep{visrag}.
For image-text retrieval, we include the MMEB training datasets~\citep{vlm2vec}, covering CIRR, NIGHTS, MSCOCO, VisualNews, N24News, ImageNet-1K, SUN397, VOC2007, HatefulMemes, A-OKVQA, OK-VQA, Visual7W, ChartQA, DocVQA, InfographicsVQA, VisDial, and WebQA.
For video tasks, the maximum sequence length is 1{,}800 tokens with 8 frames sampled per video and up to 200 tokens per frame.
For all other multimodal tasks, the maximum sequence length is 1{,}500 tokens, with up to 1{,}000 tokens allocated to the image. 
\begin{figure*}[htbp]
\centering
\includegraphics[width=0.9\linewidth]{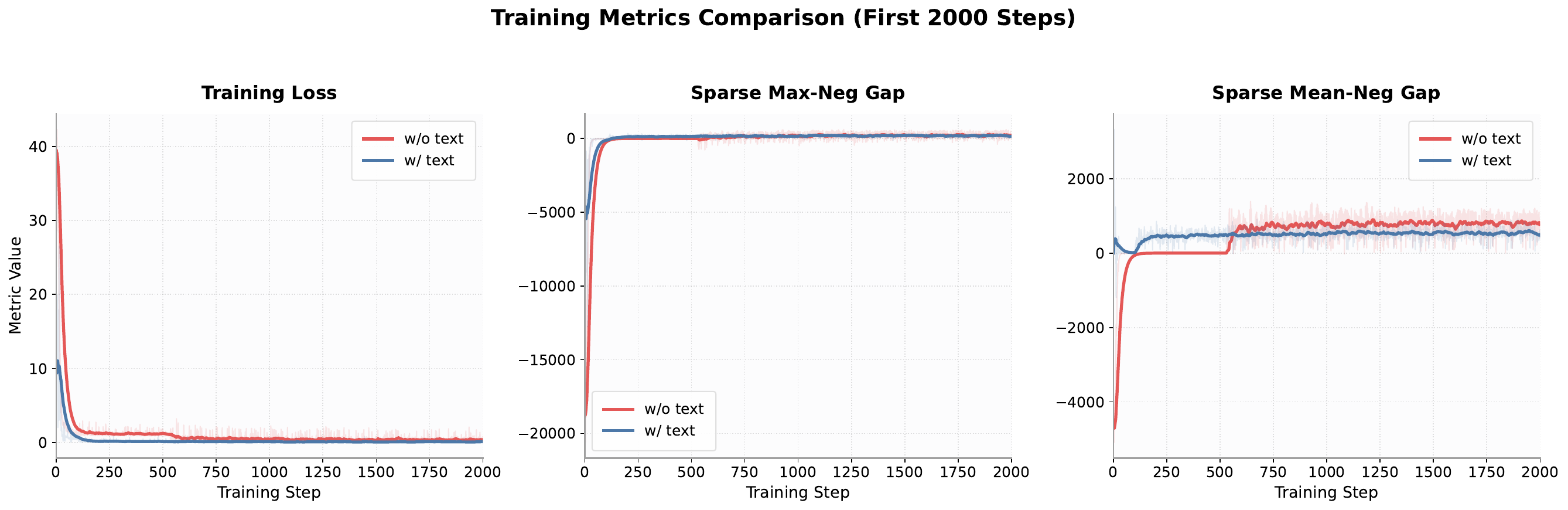}
\caption{Training dynamics comparison. We plot the loss, max negative gap, and mean negative gap during training. }
\label{fig:training-metrics}
\end{figure*}

\begin{table*}[htbp]
\centering
\begin{tabularx}{0.9\textwidth}{p{3.5cm} X}
\toprule[.1em]
\textbf{Dataset} & \textbf{Instruction} \\
\midrule
AllNLI & Given a premise, retrieve a hypothesis that is entailed by the premise. Retrieve semantically similar text. \\
DuReader & Given a Chinese search query, retrieve web passages that answer the question. \\
ELI5 & Provided a user question, retrieve the highest voted answers on Reddit ELI5 forum. \\
FEVER & Given a claim, retrieve documents that support or refute the claim. \\
HotpotQA & Given a multi-hop question, retrieve documents that can help answer the question. \\
MIRACL & Given a question, retrieve Wikipedia passages that answer the question. \\
MrTyDi & Given a question, retrieve Wikipedia passages that answer the question. \\
MSMARCO-Doc & Given a web search query, retrieve relevant documents that answer the query. \\
MSMARCO-Psg & Given a web search query, retrieve relevant passages that answer the query. \\
NQ & Given a question, retrieve Wikipedia passages that answer the question. \\
Quora & Given a question, retrieve questions that are semantically equivalent to the input question. \\
SQuAD & Retrieve Wikipedia passages that answer the question. \\
T2Ranking & Given a Chinese search query, retrieve web passages that answer the question. \\
TriviaQA & Retrieve Wikipedia passages that answer the question. \\
MLDR & Retrieve Wikipedia passages that answer the question. \\
\bottomrule[.1em]
\end{tabularx}
\caption{Retrieval instructions for text training datasets.}
\label{tab:train_instructions_text}
\end{table*}

\subsection{Training Details}

We initialize from a pretrained multimodal backbone and add $N{=}16$ special tokens to the tokenizer and embedding matrix.
Training uses LoRA~\citep{lora2022} applied to the attention and MLP projections (\texttt{q\_proj}, \texttt{k\_proj}, \texttt{v\_proj}, \texttt{up\_proj}, \texttt{down\_proj}, \texttt{gate\_proj}), with the visual encoder frozen.
We train in \texttt{bf16} mixed precision with DeepSpeed ZeRO and gradient checkpointing enabled.

\paragraph{Hyperparameters.}
We train the \ours models based on Qwen3.5. 
We use a cosine learning rate schedule with a peak learning rate of $3{\times}10^{-5}$ and a warmup ratio of 0.1. 
The dense retrieval temperature is set to $\tau{=}0.03$, and the sparse retrieval temperature is $\tau_s{=}32$. 
For the FLOPS regularizer, we set $\alpha_q = \alpha_d = 1{\times}10^{-4}$ with a linear ramp-up over the first 200 steps. 
The sparse loss factor $\lambda{=}1.0$ balances the sparse and dense InfoNCE losses equally.
We train for 1 epoch and fix the random seed to 42. 
Training is conducted on 16$\times$A100 GPUs for \ours-2B and \ours-4B with a per-device batch size of 16, and on 32$\times$A100 GPUs for \ours-9B with a per-device batch size of 8, ensuring a consistent batch size of 256 across all models.

\subsection{Training Analysis}

Since sparse scores are computed via inner products, sparse training can be unstable due to the large dynamic range of unnormalized similarities.
We find that incorporating text training data significantly stabilizes and accelerates multimodal sparse model training.
Following GVE~\citep{gve}, we monitor three key metrics during training: loss, max negative gap, and mean negative gap. The max negative gap is defined as the largest score difference between a positive and the negative within a batch, while the mean negative gap is the average score difference across all positive-negative pairs in a batch. 
As shown in~\autoref{fig:training-metrics}, when trained with text data, the model learns effective sparse representations rapidly ($<$100 steps). In contrast, training exclusively on multimodal data (\eg image, video) requires nearly 500 steps before the model can reliably distinguish positives from negatives.

\begin{table*}[htbp]
\centering
\begin{tabularx}{\textwidth}{p{3cm} X}
\toprule[.1em]
\textbf{Dataset} & \textbf{Instruction} \\
\midrule
CIRR & Given an image, find a similar everyday image with the described changes. \\
NIGHTS & Find a day-to-day image that looks similar to the provided image. \\
MSCOCO & Select the portion of the image that isolates the object. \\
MSCOCO (image-to-text) & Find an image caption describing the given everyday image. \\
VisualNews (image-to-caption) & Find a caption for the news in the given photo. \\
MSCOCO (text-to-image) & Find me an everyday image that matches the given caption. \\
VisualNews (text-to-image) & Retrieve an image of this news caption. \\
N24News & Classify the domain of the given news image. \\
ImageNet\_1K & Classify the given image. \\
SUN397 & Identify the scene shown in the image. \\
VOC2007 & Identify the object shown in the image. \\
HatefulMemes & Determine whether the given image constitutes hateful speech or not. \\
A-OKVQA & Answer the question about the given image. \\
OK-VQA & Answer the question about the given image. \\
Visual7W & Answer the question about the given image. \\
ChartQA & Answer the question about the given chart image. \\
DocVQA & Answer the question about the given document image. \\
InfographicsVQA & Answer the question about the given infographic image. \\
VisDial & Retrieve an image based on the given dialogue. \\
WebQA & Find a Wikipedia image that answers this question. \\
VISRAG-IID & Find a Wikipedia image that answers this question. \\
ViDoRe & Find a Wikipedia image that answers this question. \\
VisRAG-Synthetic & Find a document image that answers this question. \\
VideoCaption300k (text-to-video) & Find a video that matches the given caption. \\
VideoCaption300k (video-to-text) & Find a caption that describes the given video. \\
VideoQA240k & Answer the question about the given video. \\
\bottomrule[.1em]
\end{tabularx}
\caption{Retrieval instructions for multimodal training datasets.}
\label{tab:train_instructions_multimodal}
\end{table*}

\section{Experiment Details}\label{sec:appendix-experiment-details}

\subsection{Evaluated Models}
We provide the details of the baselines compared against \ours on the multimodal benchmark \mmebv (\autoref{tab:main_results}) and the text benchmark \beir (\autoref{tab:beir}) in~\autoref{appendix-model-configuration}.

\begin{table*}[htbp]
\centering
\small
\resizebox{0.90\textwidth}{!}{%
\begin{tabular}{llcll}
\toprule[.1em]
\textbf{Model} & \textbf{Release} & \textbf{Retrieval} & \textbf{Backbone} & \textbf{\#Params} \\
\midrule
\multicolumn{5}{c}{\emph{\textbf{Baselines}}} \\
\midrule
Qwen3-VL-Embedding-8B  & 2026-01 & Dense  & Qwen3-VL-8B   & 8B   \\
Qwen3-VL-Embedding-2B  & 2026-01 & Dense  & Qwen3-VL-2B   & 2B   \\
RzenEmbed-V2-7B        & 2025-10 & Dense  & Qwen2-VL-7B   & 7B   \\
Embed-RL-4B            & 2026-02 & Dense  & Qwen3-VL-4B   & 4B   \\
Embed-RL-2B            & 2026-02 & Dense  & Qwen3-VL-2B   & 2B   \\
Ops-MM-Embed-7B        & 2025-07 & Dense  & Qwen2-VL-7B   & 7B   \\
Ops-MM-Embed-2B        & 2025-07 & Dense  & Qwen2-VL-2B   & 2B   \\
UniME-7B               & 2025-10 & Dense  & Qwen2-VL-7B   & 7B   \\
UniME-2B               & 2025-10 & Dense  & Qwen2-VL-2B   & 2B   \\
GME-7B                 & 2024-12 & Dense  & Qwen2-VL-7B   & 7B   \\
VLM2Vec-V2-2.0B        & 2025-05 & Dense  & Qwen2-VL-2B   & 2B   \\
SPLADE-v3              & 2024-03 & Sparse & BERT    & 110M \\
Echo-Mistral-SPLADE    & 2024-08 & Sparse & Mistral-7B    & 7B   \\
\midrule
\multicolumn{5}{c}{\emph{\textbf{Our model}}} \\
\midrule
\ours-2B               & 2026-08 & Dense \& Sparse & Qwen3.5-2B & 2B \\
\ours-4B               & 2026-08 & Dense \& Sparse & Qwen3.5-4B & 4B \\
\ours-9B               & 2026-08 & Dense \& Sparse & Qwen3.5-9B & 9B \\
\bottomrule
\end{tabular}
}
\caption{Details of the baselines compared against \ours. Multimodal models are evaluated on \mmebv; text-only models are evaluated on the nine \beir datasets reported in~\autoref{tab:beir}. \ours produces both dense and sparse embeddings from a single checkpoint.}
\label{appendix-model-configuration}
\end{table*}

\subsection{Evaluation Settings}

\paragraph{Metrics.} For MMEB-v2~\citep{vlm2vecv2}, we use its evaluation metrics for different datasets. For BEIR, we use nDCG@10 as the evaluation metric.

\paragraph{Score Calculation.} Unless otherwise noted, dense scores are cosine similarities of the EOS-token hidden states (\ie the last content token preceding the special tokens, as defined in~\S\ref{sec:method}), and sparse scores are inner products over the partitioned vocabulary representation defined in~\S\ref{sec:method}. 
 
\subsection{Details of Ablation Study}\label{sec:appendix-ablation-study}

\paragraph{Effectiveness.} We further detail the hybrid scoring used in~\S\ref{sec:effectiveness}. Given a query $q$ and a candidate $d$, the hybrid score linearly combines the dense cosine similarity $s_{\text{dense}}(q,d)$ and the sparse inner product $s_{\text{sparse}}(q,d)$ as
\begin{equation}
s_{\text{hybrid}}(q, d) = \alpha\, s_{\text{dense}}(q, d) + \beta\, s_{\text{sparse}}(q, d).
\end{equation}
Because $s_{\text{dense}}$ and $s_{\text{sparse}}$ lie on very different scales, dense cosine similarities are bounded in $[-1, 1]$ while sparse inner products can take much larger magnitudes. We fix $\alpha = 1.0$ and tune $\beta$ per modality on a held-out split. We use $\beta = 5\mathrm{e}{-8}$ for Text, $\beta = 5\mathrm{e}{-4}$ for Image, $\beta = 1\mathrm{e}{-4}$ for Video, and $\beta = 7\mathrm{e}{-4}$ for VisDoc. The optimum for plain text is several orders of magnitude smaller because raw sparse inner products grow rapidly with the number of activated tokens, whereas the heavier multimodal inputs yield much sparser activation patterns and therefore admit relatively larger $\beta$ values.

\begin{figure}[htbp]
\centering
\includegraphics[width=0.6\linewidth]{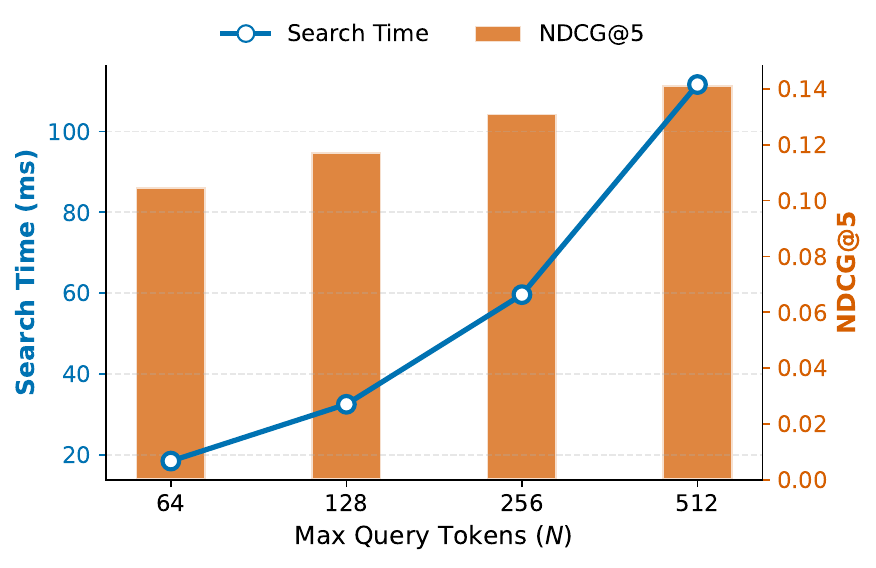}
\caption{Search efficiency--effectiveness trade-off between dense retrieval and sparse inverted-index retrieval.}
\label{fig:search-efficiency-comparison}
\end{figure}

\paragraph{Efficiency.}
The sparse representations of \ours integrate natively with inverted indices, opening the door to large-scale lexical retrieval. To quantify this, we benchmark the offline retrieval mode of BrowseComp-Plus, comparing a Faiss-based dense index against a Lucene-backed sparse inverted index.
\autoref{fig:search-efficiency-comparison} illustrates the latency--accuracy trade-off of our sparse mode: by capping the maximum number of activated tokens per query ($K_{\mathrm{prune}}$), practitioners can flexibly trade search latency for retrieval quality (NDCG@5) to match deployment constraints. While dense search yields slightly higher absolute performance at this corpus size, we expect inverted-index retrieval to become increasingly favorable as the corpus grows; a systematic multi-scale evaluation is left to future work. 

\subsection{Extra Experiment Results}

To provide a more reliable test of our hyperparameter settings, we report the MMEB-v2 results of models trained with random seed 42. These trends are broadly consistent with the main-paper settings: 
(1) joint dense-sparse training preserves dense performance while producing a competitive sparse branch ~\autoref{tab:ablation_mode}. 
(2) using too many special tokens hurts sparse retrieval ~\autoref{tab:ablation_nes}. 
(3) a higher sparse-training temperature is beneficial ~\autoref{tab:ablation_temperature}. 

\begin{table*}[htbp]
\centering
\renewcommand{\arraystretch}{1.1}
\setlength{\tabcolsep}{2.6pt}
\resizebox{\textwidth}{!}{
\begin{tabular}{l ccccc ccccc ccccc c}
\toprule
\multirow{2}{*}{\textbf{Mode}}
& \multicolumn{5}{c}{\textbf{Image}}
& \multicolumn{5}{c}{\textbf{Video}}
& \multicolumn{5}{c}{\textbf{VisDoc}}
& \multirow{2}{*}{\textbf{All}}\\
\cmidrule(lr){2-6} \cmidrule(lr){7-11} \cmidrule(lr){12-16}
& \textbf{CLS} & \textbf{QA} & \textbf{RET} & \textbf{GD} & \textbf{Avg.}
& \textbf{CLS} & \textbf{QA} & \textbf{RET} & \textbf{MRET} & \textbf{Avg.}
& \textbf{VDRv1} & \textbf{VDRv2} & \textbf{VR} & \textbf{OOD} & \textbf{Avg.} & \\
\midrule
\textbf{\# of Datasets} & 10 & 10 & 12 & 4 & 36 & 5 & 5 & 5 & 3 & 18 & 10 & 4 & 6 & 4 & 24 & 78 \\
\midrule
Dense-only & 55.3 & 65.4 & 65.3 & 83.5 & 64.6 & 56.3 & 52.2 & 44.8 & 48.3 & 50.6 & 81.6 & 61.7 & 82.9 & 66.8 & 76.1 & 64.9 \\
Sparse-only & 54.1 & 63.6 & 64.2 & 79.8 & 63.0 & 54.1 & 52.6 & 43.6 & 45.6 & 49.3 & 79.9 & 58.5 & 82.5 & 64.9 & 74.5 & 63.4 \\
UEmbed-2B (dense) & 55.5 & 65.3 & 65.1 & 83.3 & 64.5 & 58.3 & 46.9 & 45.6 & 47.5 & 49.8 & 81.6 & 59.5 & 83.1 & 66.7 & 75.8 & 64.6 \\
UEmbed-2B (sparse) & 54.7 & 63.1 & 64.8 & 81.8 & 63.4 & 56.0 & 46.3 & 44.4 & 45.7 & 48.4 & 79.9 & 59.3 & 82.0 & 65.9 & 74.7 & 63.4 \\
\bottomrule
\end{tabular}
}
\caption{MMEB-v2 performance comparison across representation modes.}
\label{tab:ablation_mode}
\end{table*}

\begin{table*}[htbp]
\centering
\renewcommand{\arraystretch}{1.1}
\setlength{\tabcolsep}{2.6pt}
\resizebox{\textwidth}{!}{
\begin{tabular}{l ccccc ccccc ccccc c}
\toprule[.1em]
\multirow{2}{*}{\textbf{$N$}}
& \multicolumn{5}{c}{\textbf{Image}}
& \multicolumn{5}{c}{\textbf{Video}}
& \multicolumn{5}{c}{\textbf{VisDoc}}
& \multirow{2}{*}{\textbf{All}}\\
\cmidrule(lr){2-6} \cmidrule(lr){7-11} \cmidrule(lr){12-16}
& \textbf{CLS} & \textbf{QA} & \textbf{RET} & \textbf{GD} & \textbf{Avg.}
& \textbf{CLS} & \textbf{QA} & \textbf{RET} & \textbf{MRET} & \textbf{Avg.}
& \textbf{VDRv1} & \textbf{VDRv2} & \textbf{VR} & \textbf{OOD} & \textbf{Avg.} & \\
\midrule
\textbf{\# of Datasets} & 10 & 10 & 12 & 4 & 36 & 5 & 5 & 5 & 3 & 18 & 10 & 4 & 6 & 4 & 24 & 78 \\
\midrule
2 & 56.0 & 62.9 & 65.0 & 81.2 & 63.7 & 56.6 & 48.2 & 45.3 & 46.6 & 49.4 & 80.1 & 59.4 & 82.1 & 65.8 & 74.8 & 63.8 \\
4 & 55.0 & 62.7 & 64.9 & 80.4 & 63.3 & 57.0 & 47.3 & 45.7 & 44.5 & 49.1 & 80.2 & 60.3 & 82.0 & 65.7 & 74.9 & 63.6 \\
8 & 54.9 & 62.2 & 65.2 & 81.0 & 63.2 & 55.1 & 48.9 & 44.8 & 46.1 & 49.0 & 79.6 & 58.8 & 81.6 & 65.8 & 74.3 & 63.4 \\
16 & 54.7 & 63.1 & 64.8 & 81.8 & 63.4 & 56.0 & 46.3 & 44.4 & 45.7 & 48.4 & 79.9 & 59.3 & 82.0 & 65.9 & 74.7 & 63.4 \\
32 & 54.6 & 62.8 & 65.0 & 80.3 & 63.2 & 54.4 & 46.2 & 41.9 & 45.7 & 47.2 & 75.6 & 22.3 & 73.4 & 65.7 & 64.5 & 59.9 \\
\bottomrule[.1em]
\end{tabular}
}
\caption{MMEB-v2 performance under different numbers of special tokens ($N$).}
\label{tab:ablation_nes}
\end{table*}

\begin{table*}[htbp]
\centering
\renewcommand{\arraystretch}{1.1}
\setlength{\tabcolsep}{2.6pt}
\resizebox{\textwidth}{!}{
\begin{tabular}{l ccccc ccccc ccccc c}
\toprule[.1em]
\multirow{2}{*}{\textbf{Temperature}}
& \multicolumn{5}{c}{\textbf{Image}}
& \multicolumn{5}{c}{\textbf{Video}}
& \multicolumn{5}{c}{\textbf{VisDoc}}
& \multirow{2}{*}{\textbf{All}}\\
\cmidrule(lr){2-6} \cmidrule(lr){7-11} \cmidrule(lr){12-16}
& \textbf{CLS} & \textbf{QA} & \textbf{RET} & \textbf{GD} & \textbf{Avg.}
& \textbf{CLS} & \textbf{QA} & \textbf{RET} & \textbf{MRET} & \textbf{Avg.}
& \textbf{VDRv1} & \textbf{VDRv2} & \textbf{VR} & \textbf{OOD} & \textbf{Avg.} & \\
\midrule
\textbf{\# of Datasets} & 10 & 10 & 12 & 4 & 36 & 5 & 5 & 5 & 3 & 18 & 10 & 4 & 6 & 4 & 24 & 78 \\
\midrule
2 & 53.7 & 61.7 & 62.8 & 77.4 & 61.6 & 50.7 & 48.6 & 41.7 & 44.3 & 46.6 & 79.5 & 57.5 & 80.0 & 64.4 & 73.4 & 61.8 \\
4 & 52.9 & 60.4 & 63.2 & 78.0 & 61.2 & 50.4 & 49.0 & 42.4 & 44.1 & 46.7 & 79.1 & 56.5 & 80.3 & 64.7 & 73.2 & 61.6 \\
8 & 55.0 & 60.8 & 64.5 & 80.2 & 62.6 & 54.9 & 50.6 & 43.2 & 43.6 & 48.6 & 80.3 & 58.1 & 81.6 & 65.3 & 74.4 & 63.0 \\
16 & 54.0 & 62.2 & 64.3 & 80.8 & 62.7 & 55.7 & 53.0 & 43.0 & 44.8 & 49.6 & 80.1 & 57.7 & 82.2 & 65.4 & 74.4 & 63.3 \\
32 & 54.7 & 63.1 & 64.8 & 81.8 & 63.4 & 56.0 & 46.3 & 44.4 & 45.7 & 48.4 & 79.9 & 59.3 & 82.0 & 65.9 & 74.7 & 63.4 \\
64 & 54.8 & 63.4 & 64.7 & 81.8 & 63.5 & 54.6 & 50.5 & 44.5 & 49.3 & 49.8 & 80.4 & 58.9 & 82.1 & 64.9 & 74.7 & 63.8 \\
\bottomrule[.1em]
\end{tabular}
}
\caption{MMEB-v2 performance under different temperature values.}
\label{tab:ablation_temperature}
\end{table*}

\section{Case Study}\label{sec:appendix-case-study}

\subsection{Qualitative Analysis}
To better understand the gap between \ours and Qwen3-VL-Embedding, we recomputed per-query Hit@1 on MMEB-v2 and inspected cases where Qwen3-VL-Embedding succeeds but UEmbed fails. The errors are modality-dependent: 
(1) Image. The gap mainly appears in fine-grained recognition (ImageNet-A, SUN397, N24News) and question-grounded VQA (GQA, ScienceQA), where UEmbed sometimes confuses visually similar categories or follows misleading lexical cues.
(2) Video. Qwen3-VL-Embedding is stronger on action recognition, moment localization, and long-video reasoning (HMDB51, UCF101, EgoSchema, NExTQA, QVHighlight), likely due to its dedicated temporal/motion supervision, while our current training contains little video-specific data.
(3) VisDoc. The gap is smaller and mostly appears in multilingual ViDoRe and table/numerical-reasoning documents. Qwen3-VL-Embedding likely benefits from its rank-KL objective, which distills reranker scores into the embedding model.

\subsection{Case Study}
We present additional case studies of \ours across different tasks and modalities. For each example, we visualize the top-10 activated sparse tokens. In the figures, \textcolor{blue}{blue} tokens denote query activations, \textcolor{green}{green} tokens denote corpus activations, and \textcolor{red}{red} tokens indicate co-activated tokens shared between the query and corpus.
We find that the sparse representations perform well on QA, Retrieval, and Grounding tasks, where the activated tokens are highly relevant to the query intent. However, for Classification, the model tends to over-associate with tangential concepts and struggles with instruction following, leading to less precise sparse activations. 

\begin{figure*}[htbp]
\centering
\includegraphics[width=\textwidth]{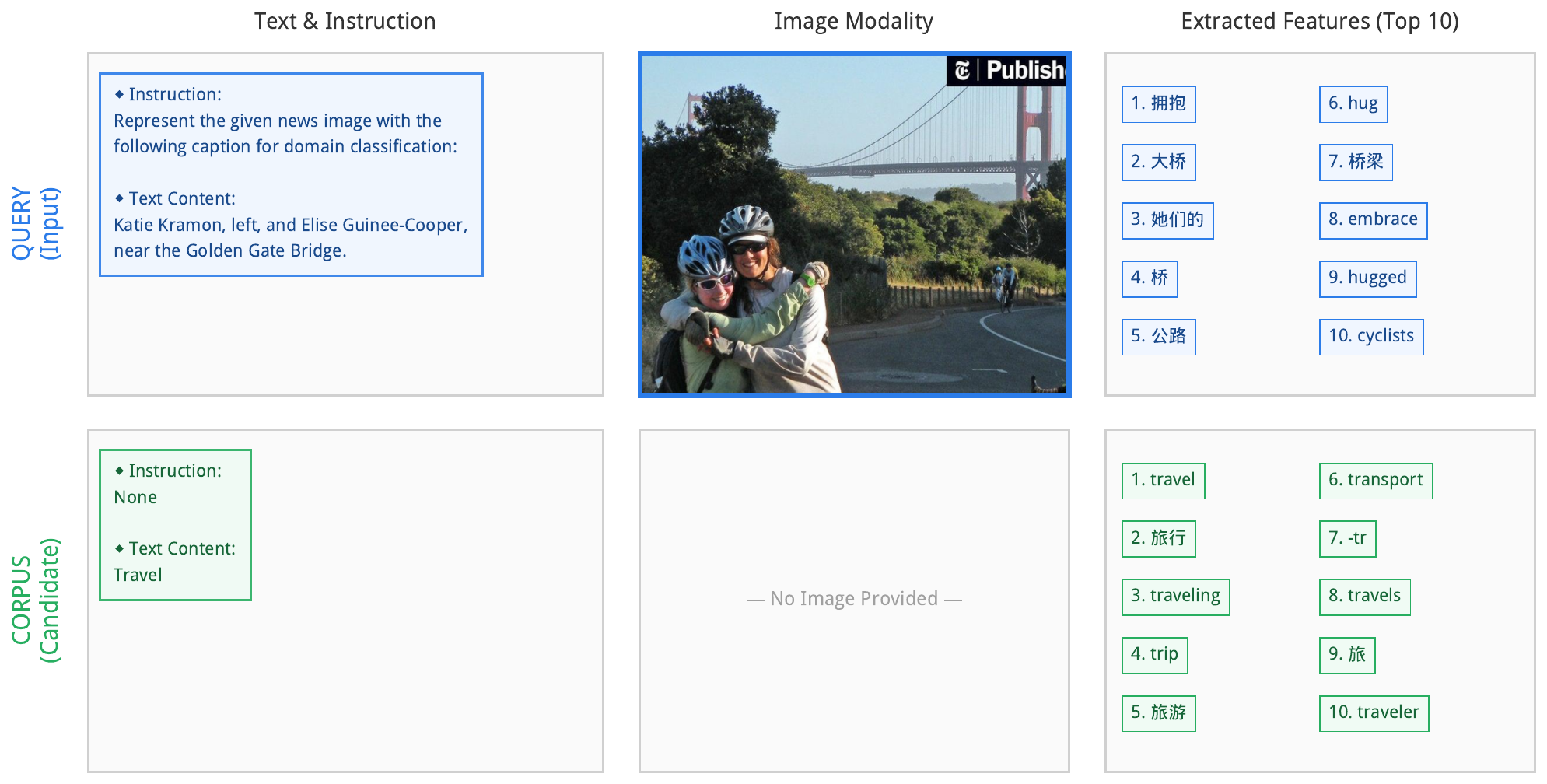}
\caption{Case study of \ours on classification (N24News). We visualize the top-10 activated sparse tokens.}
\label{fig:case-cls-n24news}
\end{figure*}

\begin{figure*}[htbp]
\centering
\includegraphics[width=\textwidth]{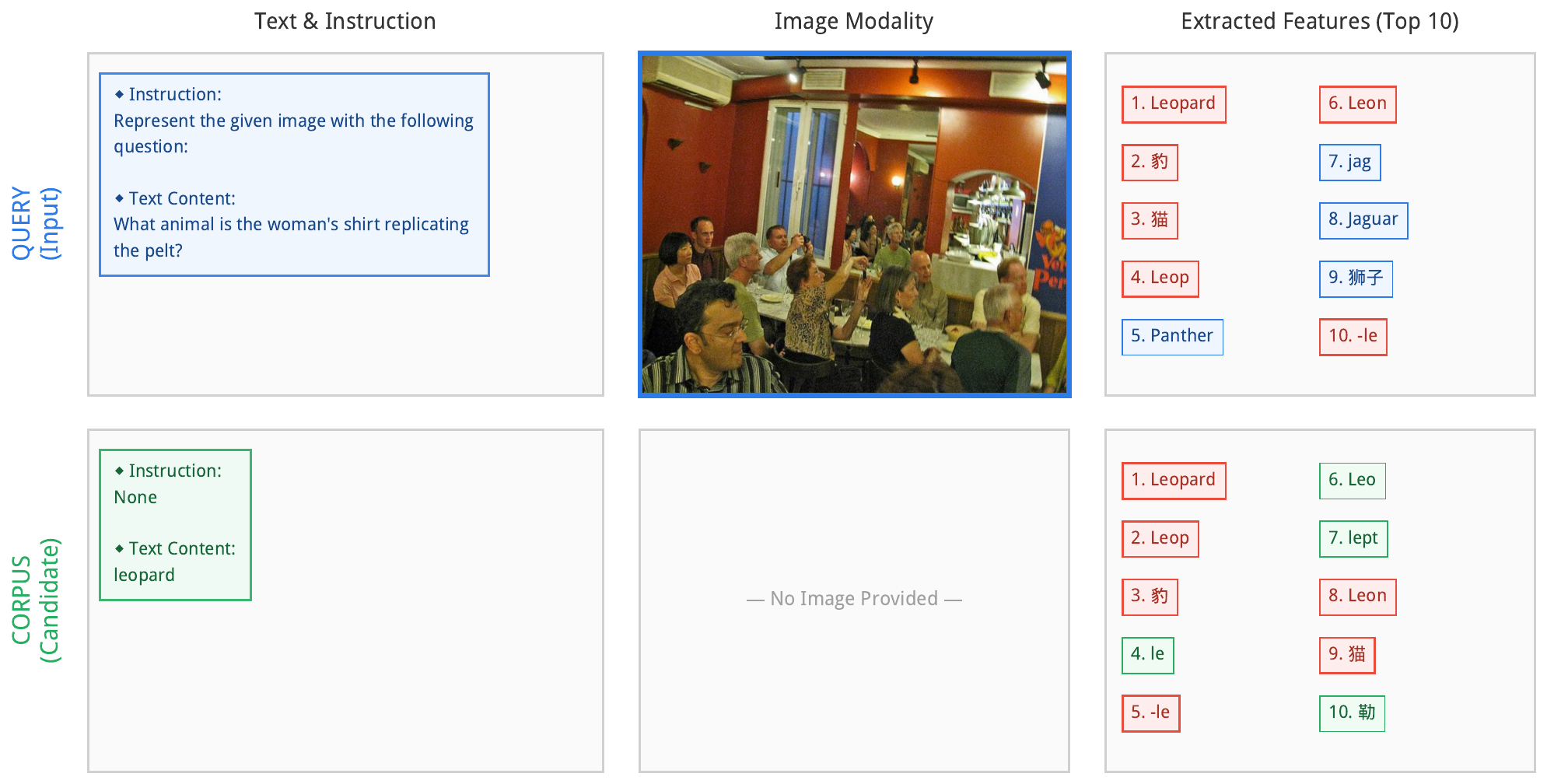}
\caption{Case study of \ours on question answering (OK-VQA). We visualize the top-10 activated sparse tokens.}
\label{fig:case-qa-okvqa}
\end{figure*}

\begin{figure*}[htbp]
\centering
\includegraphics[width=\textwidth]{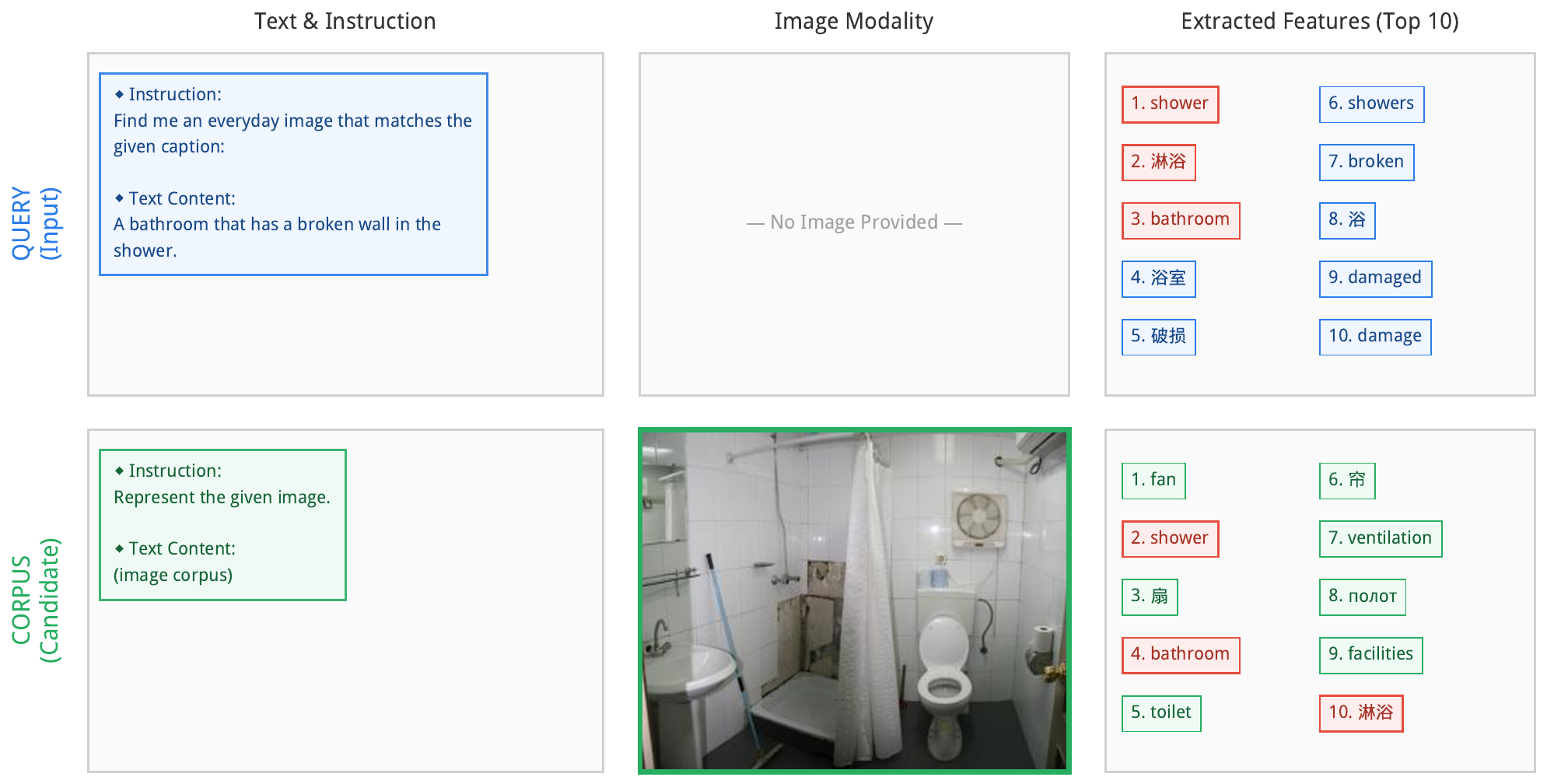}
\caption{Case study of \ours on text-to-image retrieval (MSCOCO). We visualize the top-10 activated sparse tokens.}
\label{fig:case-ret-mscoco-t2i}
\end{figure*}

\begin{figure*}[htbp]
\centering
\includegraphics[width=\textwidth]{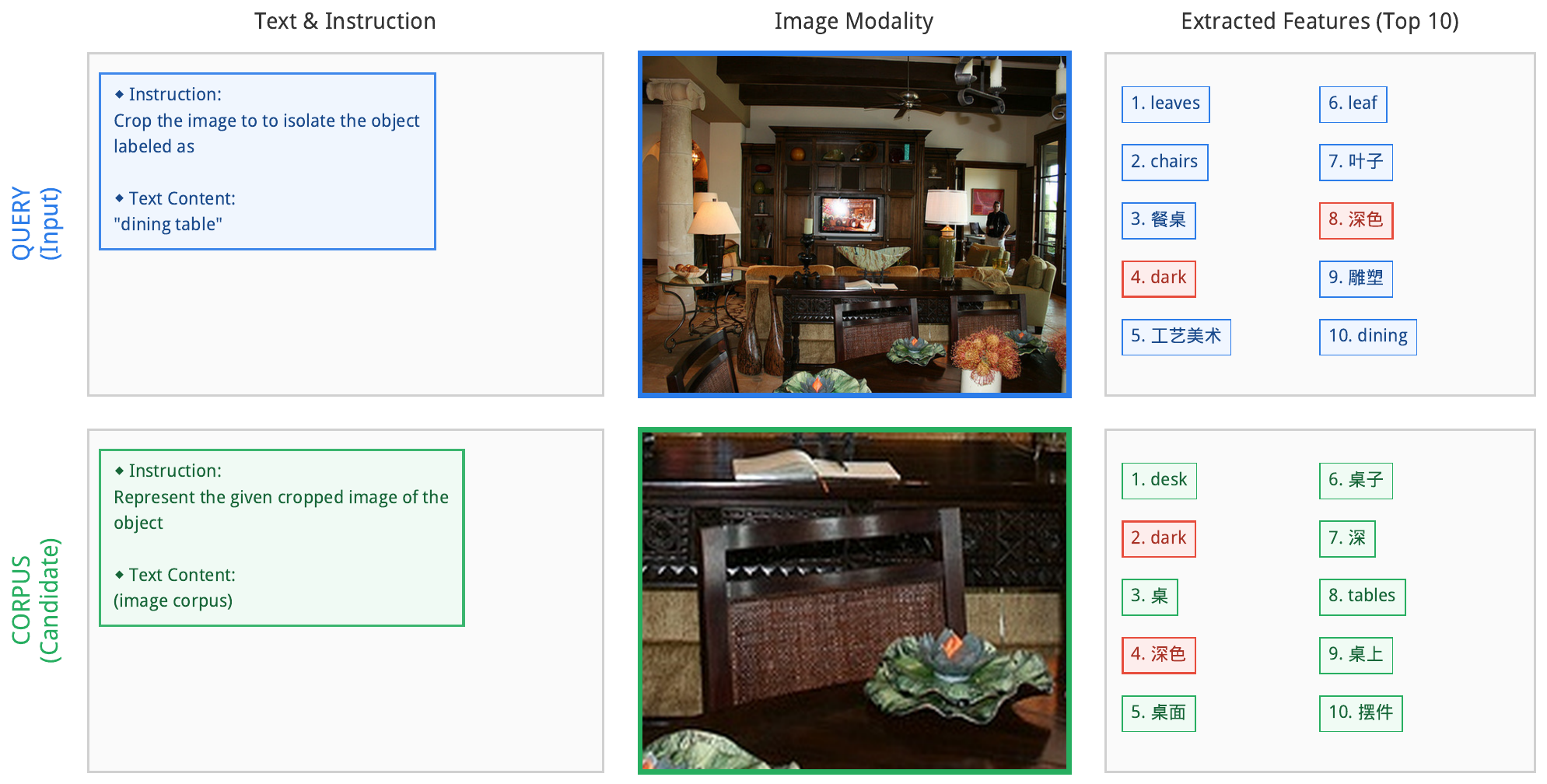}
\caption{Case study of \ours on grounding (MSCOCO). We visualize the top-10 activated sparse tokens.}
\label{fig:case-grd-mscoco}
\end{figure*}

\end{document}